\documentclass[runningheads]{llncs}
\usepackage{graphicx}

\usepackage{tikz}
\usepackage{comment}
\usepackage{amsmath,amssymb} 
\usepackage{color}
\usepackage{cases}

\usepackage[accsupp]{axessibility}  

\usepackage{bm}

\def\ie{\mbox{\textit{i.e.}, }}
\def\eg{\mbox{\textit{e.g.}, }}

\def\mL{{\mathcal L}}

\def\mP{{\mathcal P}}

\def\mT{{\mathcal T}}

\DeclareMathAlphabet\mathbfcal{OMS}{cmsy}{b}{n}

\def\0{{\bf 0}}
\def\1{{\bf 1}}

\def\bA{{\bm{A}}}

\def\bF{{\bm{F}}}

\def\bK{{\bm{K}}}

\def\bM{{\bm{M}}}

\def\bP{{\bm{P}}}
\def\bQ{{\bm{Q}}}

\def\bV{{\bm{V}}}

\def\bX{{\bm{X}}}

\def\bk{{\bm k}}

\def\bq{{\bm q}}

\def\mmE{{\mathbb E}}
\def\mmP{{\mathbb P}}

\def\mmR{{\mathbb R}}

\def\Attention{{\mathrm{Attention}}}

\def\Sigmoid{{\mathrm{Sigmoid}}}
\def\Conv{{\mathrm{Conv}}}

\def\TopK{{\mathrm{TopK}}}

\def\STL{{\mathrm{STL}}}
\def\Ref{{\mathrm{Ref}}}
\def\Tanh{{\mathrm{Tanh}}}

\usepackage{ntheorem}

\newtheorem*{*thm}{Theorem}

\newtheorem*{*lemma}{Lemma}
\usepackage{multirow}
\usepackage{subfigure}

\usepackage{algorithm}
\usepackage{algorithmic}
\usepackage{hyperref}
\usepackage{caption}
\hypersetup{colorlinks=true,urlcolor=blue}
\usepackage{wrapfig,lipsum,booktabs}
\usepackage{wrapfig, blindtext} 

\begin{document}
\pagestyle{headings}
\mainmatter
\def\ECCVSubNumber{2105}  

\title{Reference-based Image Super-Resolution with Deformable Attention Transformer} 


\titlerunning{Reference-based Image SR with Deformable Attention Transformer}
%
\author{Jiezhang Cao\inst{1} \and 
        Jingyun Liang\inst{1} \and
        Kai Zhang\inst{1} \and
        Yawei Li\inst{1} \and 
        Yulun Zhang\inst{1}\thanks{Corresponding author.} \and 
        Wenguan Wang\inst{1} \and
        Luc Van Gool\inst{1,2}
        }
\authorrunning{Jiezhang Cao et al.} 
%
\institute{$^{1}$Computer Vision Lab, ETH Zürich, Switzerland \quad~~$^{2}$KU Leuven, Belgium \\
\email{\{jiezhang.cao, jingyun.liang, kai.zhang, yawei.li, yulun.zhang, wenguan.wang, vangool\}@vision.ee.ethz.ch}\\
\url{https://github.com/caojiezhang/DATSR}
}
\maketitle

\begin{abstract}
Reference-based image super-resolution (RefSR) aims to exploit auxiliary reference (Ref) images to super-resolve low-resolution (LR) images. Recently, RefSR has been attracting great attention as it provides an alternative way to surpass single image SR. However, addressing the RefSR problem has two critical challenges: (i) It is difficult to match the correspondence between LR and Ref images when they are significantly different; (ii) How to transfer the relevant texture from Ref images to compensate the details for LR images is very challenging. To address these issues of RefSR, this paper proposes a deformable attention Transformer, namely DATSR, with multiple scales, each of which consists of a texture feature encoder (TFE) module, a reference-based deformable attention (RDA) module and a residual feature aggregation (RFA) module. Specifically, TFE first extracts image transformation (\eg brightness) insensitive features for LR and Ref images, RDA then can exploit multiple relevant textures to compensate more information for LR features, and RFA lastly aggregates LR features and relevant textures to get a more visually pleasant result. Extensive experiments demonstrate that our DATSR achieves state-of-the-art performance on benchmark datasets quantitatively and qualitatively. 
\keywords{Reference-based Image Super-Resolution, Correspondence Matching, Texture Transfer, Deformable Attention Transformer}
\end{abstract}

\section{Introduction}
Single image super-resolution (SISR), which aims at recovering a high-resolution (HR) image from a low-resolution (LR) input, is an active research topic due to its high practical values
\cite{Jo_2021_CVPR,zhang2018ffdnet,li2019SRFBN,kai2021bsrgan,Wang_2021_CVPR,Khrulkov_2021_CVPR,Xing_2021_CVPR,liang2021mutual,Kong_2021_CVPR,Kar_2021_CVPR,guo2020hierarchical,liang2021hierarchical}.
However, SISR is a highly ill-posed problem since there exist multiple HR images that can degrade to the same LR image~\cite{ulyanov2018deep,guo2020drn}.
While real LR images usually have no corresponding HR ground-truth (GT) images, one can easily find a high-quality image as a reference (Ref) image with high-frequency details from various sources, such as photo albums, video frames, and web image search, which has similar semantic information (such as content and texture) to the LR image. 
Such an alternative SISR method is referred to as reference-based super-resolution (RefSR), which aims to transfer HR textures from the Ref images to super-resolved images and has shown promising results over SISR. 
Although various RefSR methods \cite{jiang2021robust,lu2021masa,yantowards,xie2020feature} have been recently proposed, two challenges remain unsolved for SR performance improvement. 

\begin{figure}[t]
\setlength\belowcaptionskip{-5pt}
\centering
\includegraphics[width=1\linewidth]{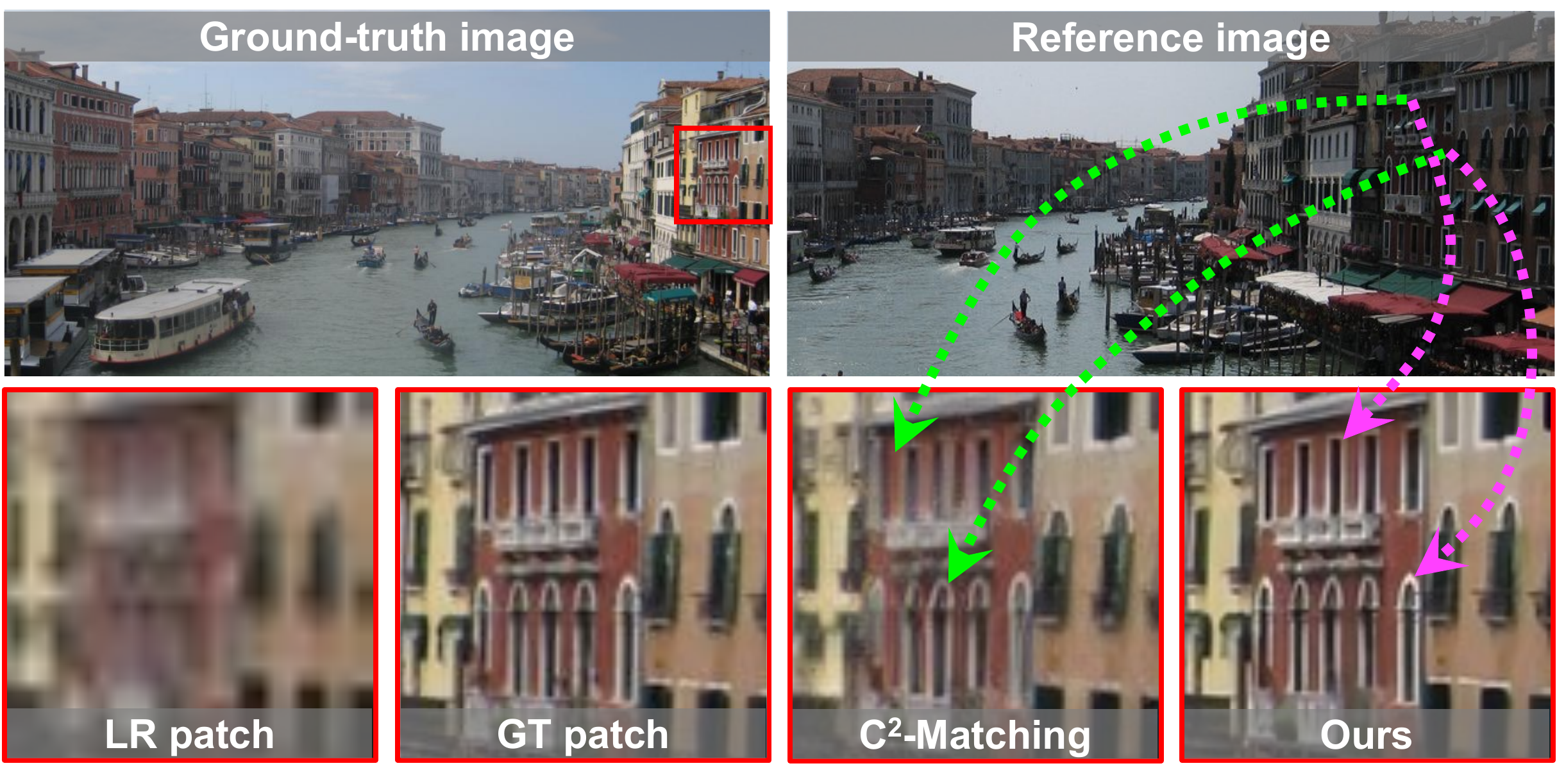}
\caption{Comparison with the state-of-the-art RefSR method $C^2$-Matching \cite{jiang2021robust}. When the brightness of LR and Ref image is different, our method performs better than $C^2$-Matching \cite{jiang2021robust} in transferring relevant textures from the Ref image to the SR image, which is closer to the ground-truth image. 
}
\label{fig:teaser}
\end{figure}

First, it is difficult to match the correspondence between the LR and Ref images especially when their distributions are different. 
For example, the brightness of the Ref images is different from that of the LR images.
Existing methods \cite{zhang2019image,yang2020learning} mostly match the correspondence by estimating the pixel or patch similarity of texture features between LR and Ref images.
However, such similarity metric is sensitive to image transformations, such as brightness and color of images.
Recently, the state-of-the-art (SOTA) method $C^2$-Matching \cite{jiang2021robust} 
trains a feature extractor, which demonstrates strong robustness to scale and rotation. 
However, it neglects to explore the effects of brightness, contrast, and color of images.
As a result, this method may transfer inaccurate textures from the Ref image, when the Ref images have different brightness from the LR image, as shown in Fig.~\ref{fig:teaser}. 
Based on the observation and analyses, we can see that the quality of correspondence is affected by the similarity metric and the distribution gap between the LR and Ref images.

On the other hand, some methods~\cite{zheng2018crossnet,Shim_2020_CVPR} adopt optical flow or deformable convolutions~\cite{dai2017deformable,zhu2019deformable,chan2021basicvsrpp,wang2019edvr} to align spatial features between the Ref and LR images.
However, these methods may find an inaccurate correspondence when the distance between the LR and Ref images is relatively large.
With the inaccurate correspondence, their performance would deteriorate seriously since the irrelevant texture cannot provide meaningful details. 
Therefore, how to accurately match the correspondence between the Ref and LR images is a challenging problem as it affects the quality of super-resolved results.

Second, it is also challenging to transfer textures of the high-quality Ref images to restore the HR images.
One representative work CrossNet \cite{zheng2018crossnet} estimates the flow from the Ref image to the LR image and then warp the features based on the optical flow.
However, the optical flow may be inaccurate, since the Ref and LR images could be significantly different.
In addition, most existing methods \cite{zhang2019image,yang2020learning,lu2021masa} search the most similar textures and the corresponding position, and then swap the texture features from the Ref image.
As a result, these methods may transfer irrelevant textures to the output and have poor SR performance, when the original estimated flow or position is inaccurate.
Hence, it is important and necessary to explore a new architecture to adaptively transfer texture and mitigate the impact of inaccurate correspondence in the Ref image.

To address the above two challenges, we propose a novel deformable attention Transformer, namely DATSR, for reference-based image super-resolution.
DATSR is built on the U-Net and consists of three basic modules, including texture feature encoders, deformable attention, and residual feature aggregation. 
Specifically, we first use texture feature encoders to extract multi-scale features with different image transformations.
Then, we propose a reference-based deformable attention to discover the multiple relevant correspondences and adaptively transfer the textures.
Last, we fuse features and reconstruct the SR images using residual feature aggregation.
We conduct extensive comparisons with recent representative SOTA methods on benchmark datasets. The quantitative and visual results demonstrate that our DATSR achieves the SOTA performance.
 
The main contributions are summarized as follows:
\begin{itemize}
\setlength{\itemsep}{5pt}
\item We propose a novel reference-based image super-resolution with deformable attention transformer (DATSR), which is end-to-end trainable by incorporating Transformer into RefSR. Compared with existing RefSR methods, our DATSR performs more robust correspondence matching and texture transfer and subsequently achieves SOTA performance quantitatively and visually. 
\item We design a new reference-based deformable attention module for correspondence matching and texture transfer.
Different from existing transformer-based methods, our transformer is built on U-Net with multi-scale features and alleviates the resolution gap between Ref and LR images.
Moreover, our transformer relieves the correspondence mismatching issue and the impact of distribution gap between LR and Ref images.
\item We conduct extensive experiments on benchmark datasets to demonstrate that our DATSR achieves SOTA performance and is also robust to different image transformations (\eg brightness, contrast and hue). 
Moreover, we find that our DATSR trained with a single Ref image outperforms existing Multi-RefSR methods trained with multiple Ref images. 
In addition, our DATSR still shows good performance even in some extreme cases, when the Ref images have no texture information.
\end{itemize}

\section{Related Work}
We will briefly introduce two related super-resolution paradigms, including single image super-resolution and reference-based image super-resolution.

\noindent{\textbf{{Single image super-resolution (SISR).}}}
The goal of SISR is to recover high-resolution (HR) images from the low-resolution (LR) images. 
Recent years have witnessed significant achievements of using deep neural networks to solve SISR \cite{dong2015image,zhang2018image}.
SRCNN \cite{dong2015image} is the pioneer work of exploiting deep convolutional networks to map LR image into HR image.
To further improve SR performance, researchers resort to employing deeper neural networks with attention mechanisms and residual blocks~\cite{lim2017enhanced,sajjadi2017enhancenet,zhang2018image,liang2021swinir,Mei_2021_CVPR,liang2021hierarchical,Zhang_2021_CVPR,Liu_2020_CVPR,liang2021mutual,zhang2021designing,Song_2020_CVPR,Dai_2019_CVPR}. 
However, it is difficult for traditional SISR methods to produce realistic images when the HR textures are highly degraded.
To relieve this, some SR methods \cite{ledig2017photo,wang2018esrgan,zhang2019ranksrgan,Hui_2021_CVPR,Wang_2020_CVPR,wang2021real,zhou2019kernel} adopt generative adversarial networks (GANs) to further improve the perceptual quality of the super-resolved outputs.

\noindent{\textbf{{Reference-based image super-resolution (RefSR).}}}
Different from SISR, RefSR has auxiliary HR images and aims to super-resolves images by transferring HR details of Ref images.
Such auxiliary information can be extracted from the reference images which are similar to HR ground-truth images.
CrossNet \cite{zheng2018crossnet} estimates the optical flow (OF) between Ref and LR images and then performs the cross-scale warping and concatenation.
Instead of estimating OF, SRNTT \cite{zhang2019image} calculates the similarity between the LR and Ref images and transfer the texture from the Ref images.
Similarly, SSEN \cite{Shim_2020_CVPR} proposes a similarity search and extraction network and it is aware of the best matching position and the relevancy of the best match.
To improve the performance, TTSR \cite{yang2020learning} proposes a hard and soft attention for texture transfer and synthesis.
Instead of using the features of a classifier, E2ENT$^2$ \cite{xie2020feature} transfers texture features by using a SR task-specific features. 
To improve the efficiency of matching, MASA \cite{lu2021masa} proposes a coarse-to-fine correspondence matching module and a spatial adaptation module to map the distribution of the Ref features to that of the LR features. 
Recently, a strong RefSR method $C^2$-Matching \cite{jiang2021robust} first proposes a contrastive correspondence network to learn correspondence, and then adopts a teacher-student correlation distillation to improve LR-HR matching, and last uses a residual feature aggregation to synthesize HR images.

It should be noted that RefSR can be extended to the case of multiple reference images, called \textbf{Multi-RefSR}, which aims to transfer the texture features from multiple Ref images to the SR image. 
Recently, a content independent multi-reference super-resolution model CIMR-SR \cite{yantowards} is proposed to transfer the HR textures from multiple reference images.
To improve the performance, AMRSR \cite{pesavento2021attention} proposes an attention-based multi-reference super-resolution network to match the most similar textures from multiple reference images.
Different from RefSR, Multi-RefSR can exploit more training information as it has multiple Ref images.
In this paper, we mainly study RefSR and train the model with single Ref image.
Nevertheless, we still compare our model with the above Multi-RefSR methods to further demonstrate the effectiveness of our DATSR.

\begin{figure*}[t]
\setlength\belowcaptionskip{-5pt}
\centering
\includegraphics[width=1\linewidth]{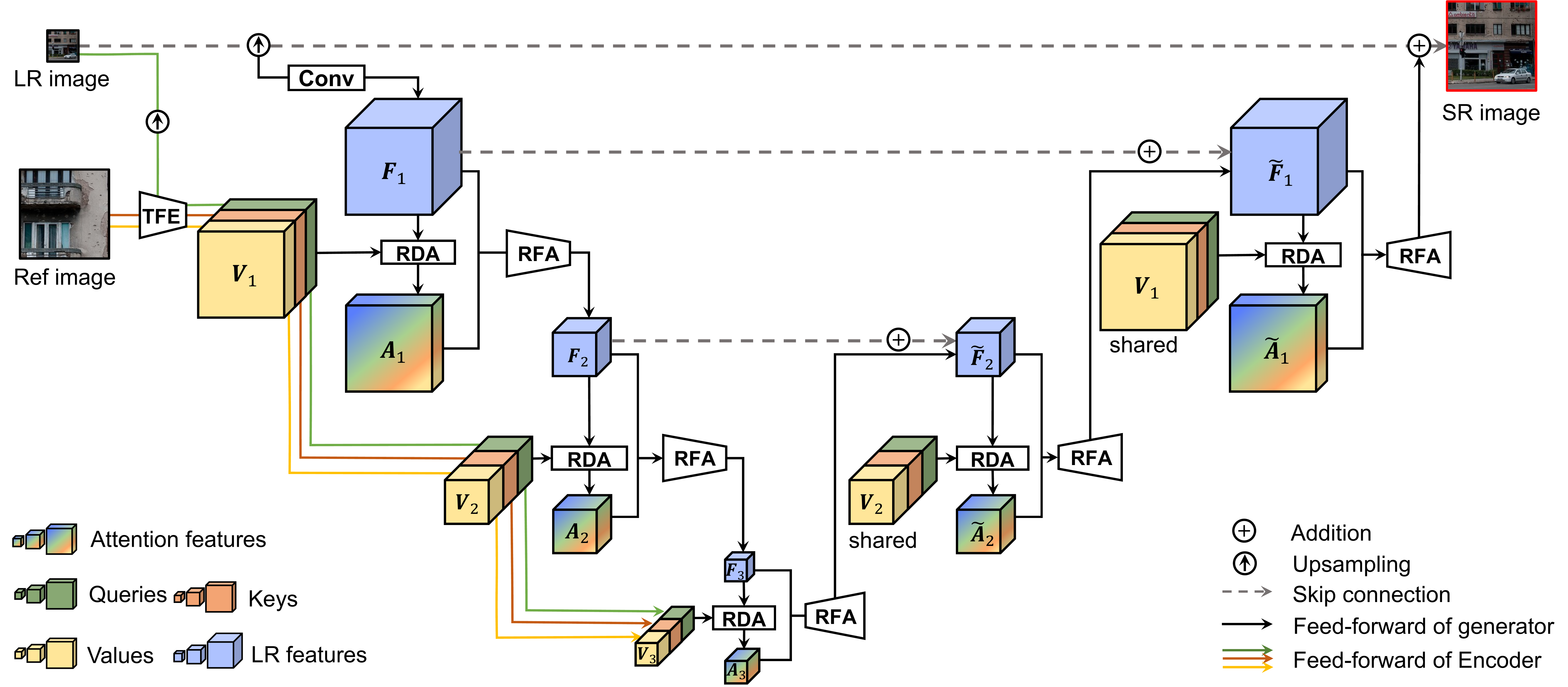}
\caption{The architecture of our DATSR network. At each scale, our model consists of texture feature encoders (TFE), a reference-based deformable attention (RDA) module and a residual feature aggregation module (RFA). }
\label{fig:framework}
\end{figure*}

\section{Proposed Method} 
Due to the the intrinsic complexity of RefSR, we divide the problem into two main sub-tasks: correspondence matching and texture transfer.
To address these, we propose a multi-scale reference-based image SR with deformable Transformer, as shown in Fig. \ref{fig:framework}.
Specifically, we first use TFE to extract multi-scale texture features of Ref and LR images, then propose RDA to match the correspondences and transfer the textures from Ref images to LR images, and last use RFA to aggregate features and generate SR images. 

\subsection{Texture Feature Encoders}
In the RefSR task, it is important to discover robust correspondence between LR and Ref images.
However, there are some underlying gaps between LR and Ref images, \ie the resolution gap and the distribution gap (\eg brightness, contrast and hue).
To address this, we propose texture feature encoders to extract robust features of LR and Ref images.
For the resolution gap, we propose to use pre-upsampling in the LR image and extract multi-scale features of LR and Ref images.
Specifically, given an LR image $\bX_{LR}$ and a reference image $\bX_{Ref}$, we upsample the LR image to the resolution of the Ref image, denoted as $\bX_{LR\uparrow}$. 
Then, we calculate multi-scale features of the LR and Ref images, \ie 
\begin{align}
    \bQ_l =\; E_{l}^q (\bX_{LR\uparrow}), \quad \bK_l =\; E_{l}^k (\bX_{Ref}), \quad \bV_l =\; E_{l}^v (\bX_{Ref}),
\end{align}
where $E_{l}^q, E_{l}^k$ and $E_l^v$ are feature encoders at the $l$-th scale. 
In our architecture, we use three scales in the texture feature encoders.
With the help of the multi-scale features in U-Net, we are able to alleviate the resolution gap between the Ref and LR images since they contain the complementary scale information. 

For the distribution gap, we augment images with different image transformations (\eg brightness, contrast and hue) in the training to improve the robustness of our model.
In addition to data augmentation, we use contrastive learning to train the encoder be less sensitive to different image transformations, inspired by \cite{jiang2021robust}.
To estimate the stable correspondence between $\bX_{LR\uparrow}$ and $\bX_{Ref}$, the feature encoders $E_{l}^q$ and $E_{l}^k$ are the same, and the feature encoder $E_l^r$ is pre-trained and fixed in the training.
In contrast, TTSR \cite{yang2020learning} directly uses a learnable feature encoder, resulting in limited performance since the textures are changing during training and the correspondence matching is unstable.
For $C^2$-Matching \cite{jiang2021robust}, it neglects to improve the robustness to brightness, contrast and hue.
To address these, we propose to learn robust multi-scale features $\bQ_l, \bK_l, \bV_l$, which can be regraded as Query, Key, and Value, and can be used in our attention mechanism conditioned on the LR features. 

\subsection{Reference-based Deformable Attention} 

\begin{figure}[t]
\centering
\begin{minipage}[t]{0.51\textwidth}
\centering
\includegraphics[width=6.0cm,trim=0 0 0 0]{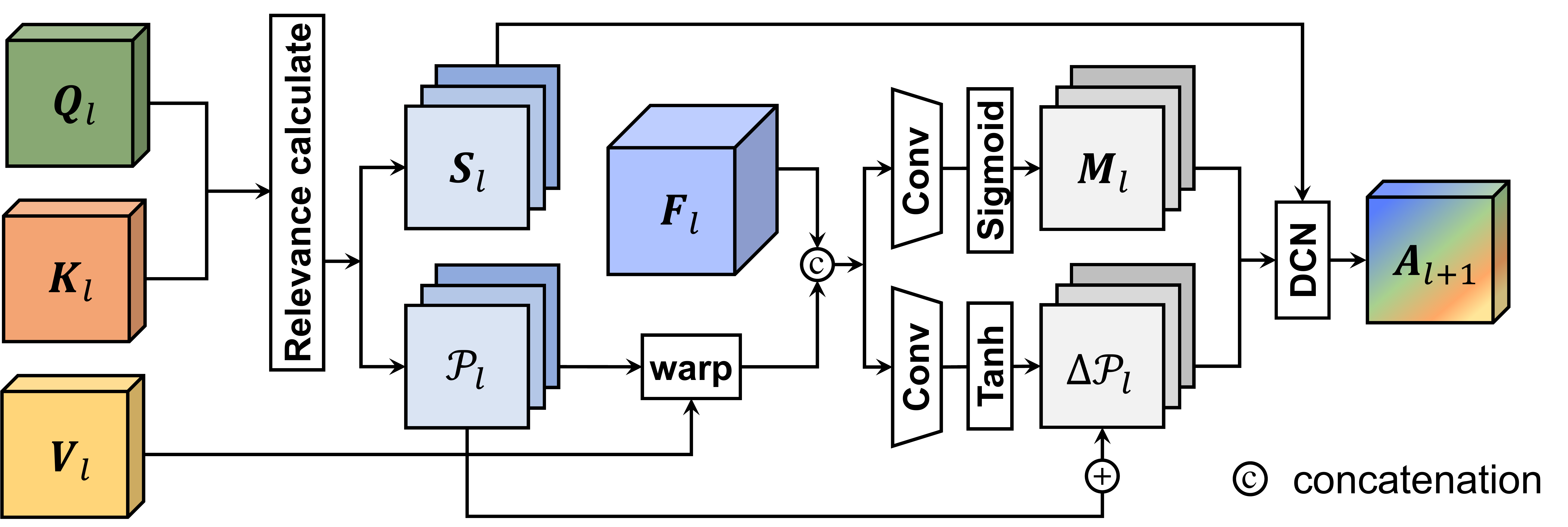}
\caption{The architecture of RDA.}
\label{fig:rda}
\end{minipage}
\begin{minipage}[t]{0.44\textwidth}
\centering
\includegraphics[width=5.5cm,trim=0 0 0 0]{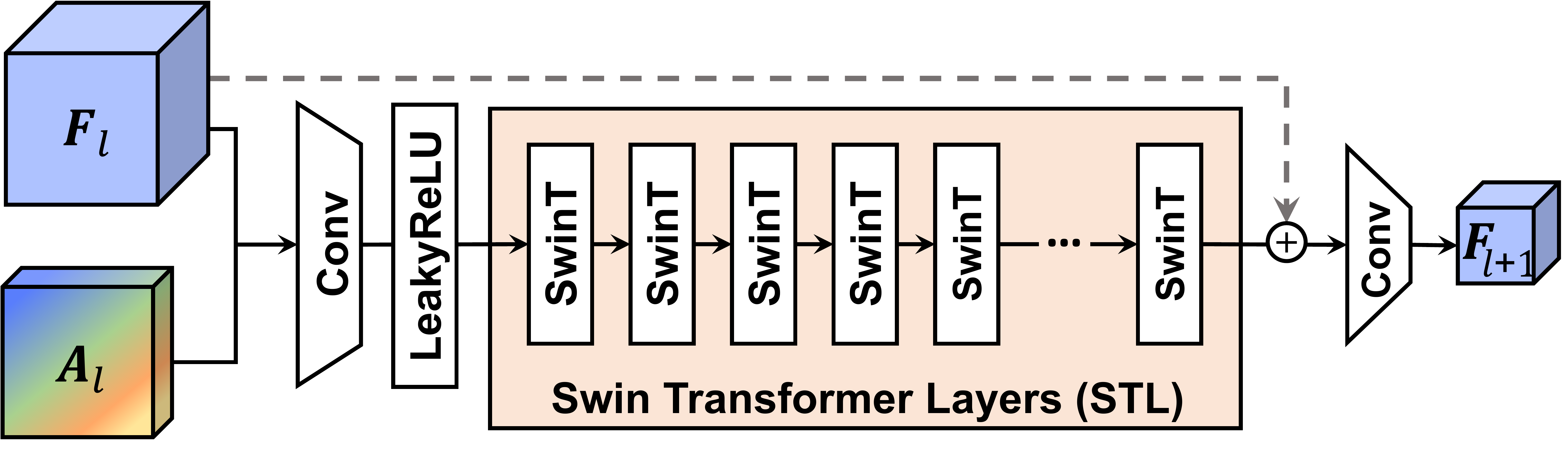}
\caption{The architecture of RFA.}
\label{fig:rfa}
\end{minipage}
\end{figure}

Existing attention-based RefSR methods (\eg \cite{yang2020learning}) tend to suffer from limited performance when the most relevant features between LR and Ref images are inaccurate, \ie the learned LR features may not well match the Ref features. 
To address this, we propose a new reference-based attention mechanism, called RefAttention, as shown in Fig. \ref{fig:rda}.
Formally, given Query $\bQ_l$, Key $\bK_l$, Value $\bV_l$, and LR features $\bF_l$, the attention feature $\bA_l$ is defined as follows:
\begin{equation}
\begin{aligned}
    \bA_l = \; \Ref\Attention(\bQ_l, \bK_l, \bV_l, \bF_l) 
    =\; \mT \left( \sigma \left( \bQ^{\top}_l\! \bK_l \right), \bV_l, \bF_l \right).
\end{aligned}
\end{equation}
Different from existing attention mechanism \cite{vaswani2017attention}, our attention is conditioned on the LR features and designed for the RefSR task.
In Fig. \ref{fig:rda}, we denoted by $\bA_l$ and $\bF_l$ in the downscaling process, and $\tilde{\bA}_l$ and $\tilde{\bF}_l$ in the upscaling process.
$\sigma(\cdot)$ is a correspondence matching function to calculate the relevance between the Ref and LR images.
Based on the relevance, we propose a texture transfer function $\mT(\cdot)$ to transfer the textures from the Ref to the LR image.

\paragraph{\textbf{\emph{Correspondence matching.}}}
The first important sub-task in RefSR is to match correspondences between LR and Ref images.
Most existing methods \cite{zhang2019image,yang2020learning} are sensitive to different image transformations (\eg brightness, contrast and hue) and may match inaccurate correspondences.
To relieve this issue, we propose a correspondence matching module in our $\Ref\Attention$, as shown in Fig. \ref{fig:rda}.
Specifically, we estimate the relevance between $\bX_{LR\uparrow}$ and $\bX_{Ref}$ by calculating similarity between $\bQ_l \in \mmR^{C{\times} H_1{\times}W_1}$ and $\bK_l \in \mmR^{C{\times} H_2{\times}W_2}$.
First, we unfold $\bQ_l$ and $\bK_l$ into patches $\bQ'_l=[\bq_1, \ldots, \bq_{H_1 W_1}]{\in}\mmR^{C {\times} H_1 W_1}$ and $\bK'_l =[\bk_1, \ldots, \bk_{H_2 W_2}] {\in} \mmR^{C {\times} H_2 W_2}$.
Then, for the given query $\bq_i$ in $\bQ'$, the top $K$ relevant positions in $\bK'$ can be calculated by normalized inner product, 
\begin{align}\label{eqn:topK}
    \bP_i = \left[ \sigma \left( \bQ'^{\top}_l \bK'_l \right) \right]_{i} = \TopK_j \left( \widetilde{\bq}_i \cdot \widetilde{\bk}_j \right), 
\end{align}
where $\widetilde{\bq}_i={\bq_i}/{\|\bq_i\|}$ and $\widetilde{\bk}_j={\bk_j}/{\|\bk_j\|}$ are normalized features, and $\TopK(\cdot)$ is a function and returns top $K$ relevant positions $\bP_i{=}\left\{p_i^1, \ldots, p_i^K\right\}$. 
Here, $\bP_i$ is the $i$-th element of $\mP_l$, and the position $p_i^1$ is the most relevant position in the Ref image to the $i$-th position in LR.
When $K>1$, it helps discover multiple correspondences, motivated by KNN \cite{Liu_2015_CVPR}.
For fair comparisons with other RefSR methods, we set $K=1$ and exploit the most relevant position in the experiments.

\paragraph{\textbf{\emph{Similarity-aware texture transfer.}}}
The second important sub-task in RefSR is to transfer textures from Ref images to LR images based on the matched correspondence.
Most existing RefSR methods \cite{zhang2019image,yang2020learning} directly swap the most relevant texture from Ref image.
However, it may degrade the performance when the most relevant texture is inaccurate. 
To address this, we propose to improve the deformable convolution (DCN) \cite{dai2017deformable,zhu2019deformable} to transfer the texture around every position $p_i^k$ of Ref images. 
Specifically, let $\Delta p_i^k$ be the spatial difference between the position $p_i$ and the $k$-th relevant position $p_i^k$, \ie $\Delta p_i^k = p_i^k - p_i$.
Then, we calculate a feature at the position $p$ using modified DCN, \ie
\begin{align}
    \bA_l(p_i) = \sum\nolimits_{k=1}^K s_i^k \sum\nolimits_{j} w_j \bV_l(p_i + \Delta p_i^k + p_j + \Delta p_j)\; m_j,
\end{align}
where $p_j \in \{ (-1, 1), (-1, 0), \ldots, (1, 1) \}$, $s_i^k$ is the cooperative weight to aggregate the $K$ textures from the Ref image, \ie $s_i^k = {\exp({ \widetilde{\bq}_i \cdot \widetilde{\bk}_{p_i^k} })} / {\sum\nolimits_{j \in \bP_i} \exp ({\widetilde{\bq}_i \cdot \widetilde{\bk}_j })} $, $w_j$ is the convolution kernel weight, $\Delta p_j$ is the $j$-th learnable offset of $\Delta \mP_l$, and $m_j$ is the $j$-th learnable mask of $\bM_l$, which can be calculated as follows,
\begin{equation}
\left\{
\begin{aligned}
    \Delta \mP_l =\;& r \cdot \Tanh\left(\Conv([\bF_l;\; \omega(\bV_l, \mP_l)])\right), \\
    \bM_l =\;& \Sigmoid \left(\Conv([\bF_l;\; \omega(\bV_l, \mP_l)]) \right),
\end{aligned}
\right.
\end{equation}
where $\omega$ is a warping function, $[;]$ is a concatenation operation, $\Conv$ is convolutional layers.
$\Sigmoid$ and $\Tanh$ are activation functions, $r$ is the max magnitude which is set as 10 in default, and $\bF_l$ is the feature of upsampled LR images at the $l$-th scale. 
With the help of the mask, we can adaptively transfer textures even if LR and Ref images are significantly different.  
When the Ref image has irrelevant texture or no information, our model is able to guild whether to transfer the textures in Ref images. 
In this sense, it can relieve the correspondence mismatching issue. 
In this paper, we mainly compare with RefSR methods with single Ref image.
Thus, we transfer one relevant textures from the Ref image for fair comparison.
With the help of our architecture, the proposed RDA module is able to improve the RefSR performance by transferring textures at each scale in both downscaling and upscaling, which is different from $C^2$-Matching \cite{jiang2021robust}.

\subsection{Residual Feature Aggregation}
To aggregate the multi-scale LR features at different layers and the transferred texture features, we propose a residual feature aggregation module (RFA) to perform feature fusion and extraction.
As shown in Fig. \ref{fig:rfa}, RFA consists of CNNs and Swin Transformer layers (STL) \cite{liu2021swin} which gain much attention in many tasks \cite{liang2021swinir,cao2021swin,liu2021video}.
Specifically, we first use a convolution layer to fuse the LR feature $\bF_{l}$ and attention features $\bA_l$, \ie
$\bF'_{l+1} =\; \Conv (\bF_{l}, \bA_l)$,
where $\Conv$ is convolutional layers. 
Then, we use Swin Transformer and a residual connection to extract deeper features of the LR and transferred features,
\begin{align}
    \bF'_{l+1} = \STL (\bF'_{l+1}) + \bF_l,
\end{align}
where the details of $\STL$ are put in the supplementary materials.
At the end of RFA, we use another convolutional layer to extract the features of STL, 
$ \bF_{l+1} = \Conv(\bF'_{l+1})$.  
Based on the aggregated features $\bF_{L}$ at the last scale, we synthesize SR images with a skip connection as 
\begin{align}
    \bX_{SR} = \bF_{L} + \bX_{LR\uparrow}.
\end{align}

\subsection{Loss Function}
In the training, we aim to i) preserve the spatial structure and semantic information of LR images; ii) discover more texture information of Ref images; iii) synthesize realistic SR images with high quality.
To this end, we use a reconstruction loss, a perceptual loss and an adversarial loss, which is the same as \cite{yang2020learning,jiang2021robust}.
The overall loss with the hype-parameters $\lambda_1$ and $\lambda_2$ is written as:
\begin{align}
    \mL = \mL_{rec} + \lambda_{1} \mL_{per} + \lambda_{2} \mL_{adv}.
\end{align}

\noindent\textbf{Reconstruction loss.}
In order to make the SR image $\bX_{SR}$ to be close to the HR ground-truth image $\bX_{HR}$, we adopt the following reconstruction loss
\begin{align}
    \mL_{rec} = \|\bX_{HR} - \bX_{SR} \|_1,
\end{align}
where $\|\cdot\|_1$ is the $\ell_1$-norm.

\noindent\textbf{Perceptual loss.}
To enhance the visual quality of SR images, the perceptual loss is widely used in SR models \cite{zhang2019image,jiang2021robust}.
The perceptual loss is defined as:
\begin{align}
    \mL_{per} = \frac{1}{V} \sum\nolimits_{i=1}^C \left\| \phi_i(\bX_{HR}) - \phi_i(\bX_{SR}) \right\|_F,
\end{align}
where $\|{\cdot}\|_F$ is the Frobenius norm, and $V$ and $C$ are the volume and channel number of the feature maps, respectively.
The function $\phi_i$ is the $i$-th intermediate layer in VGG19 \cite{simonyan2014very}, and we use the relu5\_1 layer of VGG19 in the experiment.

\noindent{\textbf{{Adversarial loss.}}}
To improve the visual quality of SR images, many SR methods \cite{ledig2017photo,wang2018esrgan} introduce GANs \cite{goodfellow2014GAN,arjovsky2017wasserstein} which have achieved good performance for SR.
Specifically, we use WGAN \cite{arjovsky2017wasserstein} loss as follows,
\begin{equation}
\begin{aligned}
    \mL_{adv} {=} \mathop{\mmE}\nolimits_{{\bX}_{SR} \sim \mmP_{SR}} [D({\bX}_{SR})] - \mathop{\mmE}\nolimits_{\bX_{H\!R} {\sim} \mmP_{H\!R}} [D(\bX_{HR})],
\end{aligned}
\label{eqn:adv_loss}
\end{equation}
where $D(\cdot)$ is a discriminator, $\mmP_{SR}$ is the distribution of the generated SR images, and $\mmP_{H\!R}$ is the distribution of the real data.

\section{Experiments}

\noindent\textbf{Datasets.}
In the experiment, we consider the RefSR dataset, \ie CUFED5 \cite{zhang2019image}, which consists of a training set and a testing set.
The CUFED5 training set contains 11,871 training pairs, and each pair has an original HR image and a corresponding Ref image at the size of 160${\times}$160. 
The CUFED5 testing set has 126 input images and each image has 4 reference images with different similarity levels.
For fair comparisons, all models are trained on the training set of CUFED5. 
To evaluate the generalization ability, we test our model on the CUFED5 testing set, Urban100 \cite{huang2015single}, Manga109 \cite{matsui2017sketch}, Sun80 \cite{sun2012super} and WR-SR \cite{jiang2021robust}.
The Sun80 and WR-SR datasets contain 80 natural images, and each paired with one or more reference images.  
For the Urban100 dataset, we concatenate the LR and random sampled HR images as the reference images.
For the Manga109 dataset, we randomly sample HR images as the reference images since there are no the reference images. 
All experiments are conducted for $4\times$ SR.

\paragraph{\textbf{\emph{Evaluation metrics.}}}
Existing RefSR methods \cite{yang2020learning,jiang2021robust,yang2020learning} mainly use PSNR and SSIM to compare the performance.
Here, PSNR and SSIM are calculated on the Y channel of YCbCr color space.
In general, larger PSNR and SSIM correspond to better performance of the RefSR method.
In addition, we compare the model size (\ie the number of trainable parameters) of different models.

\paragraph{\textbf{\emph{Implementation details.}}}
The input LR images are generated by bicubicly downsampling the HR images with scale factor 4.
For the encoders and discriminator, we adopt the same architectures as \cite{jiang2021robust}.
We use a pre-trained relu1\_1, relu2\_1 and relu3\_1 of VGG19 to extract multi-scale features.
we augment the training data with randomly horizontal and vertical flipping or different random rotations of 90$^{\circ}$, 180$^{\circ}$ and 270$^{\circ}$.
Besides, we also augment the training data by randomly changing different brightness, contrast and hue of an image by using ColorJitter in pytorch.
In the training, we set the batch size as 9, \ie each batch has 9 LR, HR and Ref patches.
The size of LR images is $40{\times}40$, and the size of HR and Ref images is $160{\times}160$.
Following the training of \cite{jiang2021robust}, we set the hype-parameters $\lambda_1$ and $\lambda_2$ as 1$\times10^{-4}$ and 1$\times10^{-6}$, respectively.
We set the learning rate of the SR model and discriminator as 1$\times10^{-4}$.
For the Adam optimizer, we set $\beta_1{=}0.9$ and $\beta_2{=}0.999$.
We provide more detailed network architectures and training details in the supplementary material.

\begin{table*}[t]
\setlength\belowcaptionskip{-6pt}
\setlength\abovecaptionskip{-9pt}
\caption{Quantitative comparisons (PSNR and SSIM) of SR models trained with only reconstruction loss (with the suffix `-rec').  
We group methods by SISR and RefSR. We mark the best results \textbf{in bold}. }
\begin{center}
\resizebox{1\textwidth}{!}{
\begin{tabular}{|c|l|c|c|c|c|c|c|c|c|c|c|}
  \hline
  \multirow{2}{*}{SR paradigms} & \multirow{2}{*}{Methods}        & \multicolumn{2}{c|}{CUFED5 \cite{zhang2019image}}            & \multicolumn{2}{c|}{Urban100 \cite{huang2015single}}  & \multicolumn{2}{c|}{Manga109 \cite{matsui2017sketch}}   & \multicolumn{2}{c|}{Sun80 \cite{sun2012super}}   & \multicolumn{2}{c|}{WR-SR \cite{jiang2021robust}}    \\ 
  \cline{3-12}
  &  & PSNR & SSIM & PSNR & SSIM & PSNR & SSIM & PSNR & SSIM & PSNR & SSIM \\
  \hline
  \hline
  \multirow{5}{*}{SISR}   
  & SRCNN \cite{dong2015image}         & 25.33 & 0.745    & 24.41 & 0.738   & 27.12 & 0.850   & 28.26 & 0.781   & 27.27 & 0.767   \\
  & EDSR \cite{lim2017enhanced}        & 25.93 & 0.777    & 25.51 & 0.783   & 28.93 & 0.891   & 28.52 & 0.792   & 28.07 & 0.793   \\
  & ENet \cite{sajjadi2017enhancenet}  & 24.24 & 0.695    & 23.63 & 0.711   & 25.25 & 0.802   & 26.24 & 0.702   & 25.47 & 0.699   \\
  & RCAN \cite{zhang2018image}         & 26.06 & 0.769    & 25.42 & 0.768   & 29.38 & 0.895   & 29.86 & 0.810   & 28.25 & 0.799   \\
  & SwinIR \cite{liang2021swinir}      & 26.62 & 0.790    & 26.26 & 0.797   & 30.05 & 0.910   & 30.11 & 0.817   & 28.06 & 0.797   \\
  \hline
  \hline
  \multirow{9}{*}{RefSR} & 
  CrossNet \cite{zheng2018crossnet}    & 25.48 & 0.764    & 25.11 & 0.764   & 23.36 & 0.741   & 28.52 & 0.793   & - & -   \\
  & SRNTT-rec \cite{zhang2019image}   & 26.24 & 0.784    & 25.50 & 0.783   & 28.95 & 0.885   & 28.54 & 0.793   & 27.59 & 0.780     \\
  & TTSR-rec \cite{yang2020learning}  & 27.09 & 0.804    & 25.87 & 0.784   & 30.09 & 0.907   & 30.02 & 0.814   & 27.97 & 0.792   \\
  & SSEN-rec \cite{Shim_2020_CVPR}    & 26.78 & 0.791    & - & -              & - & -              & - & -              & - & -  \\
  & E2ENT$^{2}$-rec \cite{xie2020feature} & 24.24 & 0.724    & - & -              & - & -              & 28.50 & 0.789   & - & -       \\
  & MASA-rec \cite{lu2021masa}     & 27.54 & 0.814    & 26.09 & 0.786  & 30.24 & 0.909   & 30.15 & 0.815   & 28.19 & 0.796 \\
  & $C^2$-Matching-rec \cite{jiang2021robust} & 28.24 & 0.841    & 26.03 & 0.785   & 30.47 & 0.911   & 30.18 & 0.817   & 28.32 & 0.801 \\
   \cline{2-12}
   & \textbf{DATSR-rec (Ours)}   & \textbf{28.72} & \textbf{0.856}   & \textbf{26.52} & \textbf{0.798}      & \textbf{30.49} & \textbf{0.912} & \textbf{30.20} & \textbf{0.818}   & \textbf{28.34} & \textbf{0.805} \\
\hline
\end{tabular}
}
\end{center}
\label{tab:quan_comp_rec}
\end{table*}

\begin{table*}[!t]
\setlength\belowcaptionskip{-18pt}
\setlength\abovecaptionskip{-10pt}
\caption{Quantitative comparisons (PSNR and SSIM) of SR models trained with all losses. We mark the best results \textbf{in bold}. }  
\begin{center}
\resizebox{1\textwidth}{!}{
\begin{tabular}{|c|l|c|c|c|c|c|c|c|c|c|c|}
  \hline
  \multirow{2}{*}{SR paradigms} & \multirow{2}{*}{Methods}        & \multicolumn{2}{c|}{CUFED5 \cite{zhang2019image}}            & \multicolumn{2}{c|}{Urban100 \cite{huang2015single}}  & \multicolumn{2}{c|}{Manga109 \cite{matsui2017sketch}}   & \multicolumn{2}{c|}{Sun80 \cite{sun2012super}}   & \multicolumn{2}{c|}{WR-SR \cite{jiang2021robust}}    \\ 
  \cline{3-12}
  &  & PSNR & SSIM & PSNR & SSIM & PSNR & SSIM & PSNR & SSIM & PSNR & SSIM \\
  \hline
  \hline
  \multirow{3}{*}{SISR} 
  & SRGAN \cite{ledig2017photo}          & 24.40 & 0.702   & 24.07 & 0.729   & 25.12 & 0.802   & 26.76 & 0.725    & 26.21 & 0.728   \\
  & ESRGAN \cite{wang2018esrgan}         & 21.90 & 0.633   & 20.91 & 0.620   & 23.53 & 0.797   & 24.18 & 0.651    & 26.07 & 0.726    \\
  & RankSRGAN \cite{zhang2019ranksrgan}  & 22.31 & 0.635   &  21.47 & 0.624  & 25.04 & 0.803   & 25.60 & 0.667    & 26.15 & 0.719      \\ \hline
  \hline
  \multirow{7}{*}{RefSR} 
  & SRNTT \cite{zhang2019image}      & 25.61 & 0.764   & 25.09 & 0.774   & 27.54 & 0.862   & 27.59 & 0.756    & 26.53 & 0.745       \\
  & TTSR \cite{yang2020learning}    & 25.53 & 0.765   & 24.62 & 0.747   & 28.70 & 0.886   & 28.59 & 0.774    & 26.83 & 0.762  \\
  & SSEN \cite{Shim_2020_CVPR}      & 25.35 & 0.742   & - & -              & - & -              & - & -               & - & -  \\
  & E2ENT$^{2}$ \cite{xie2020feature}   & 24.01 & 0.705   & - & -              & -  & -             & 28.13 & 0.765    & - & - \\
  & MASA \cite{lu2021masa}       & 24.92 & 0.729   & 23.78 & 0.712   & 27.26 & 0.847   & 27.12 & 0.708    & 25.74 & 0.717 \\
  & $C^{2}$-Matching \cite{jiang2021robust}~~~~~~ & 27.16 & 0.805   & 25.52 & 0.764   & 29.73 & 0.893   & 29.75 & 0.799    & 27.80 & 0.780   \\
   \cline{2-12}
  & \textbf{DATSR (Ours)}   & \textbf{27.95} & \textbf{0.835}   & \textbf{25.92} & \textbf{0.775}   & \textbf{29.75} & \textbf{0.893}   & \textbf{29.77} & \textbf{0.800}    & \textbf{27.87} & \textbf{0.787}  \\
\hline
\end{tabular}
}
\end{center}
\label{tab:quan_comp_gan}
\end{table*}

\begin{figure*}[t]
\centering
\includegraphics[width=1\linewidth]{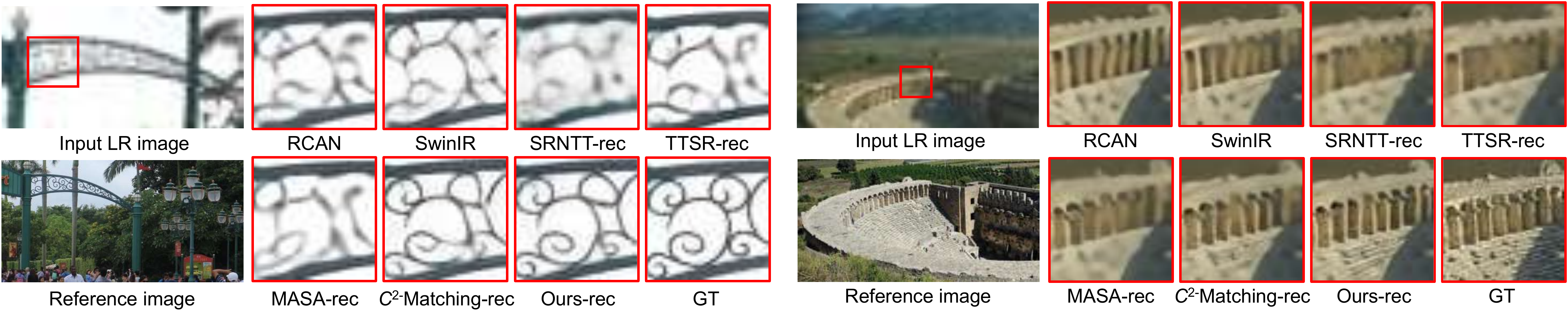}
\includegraphics[width=1\linewidth,trim=0 40 0 0 ]{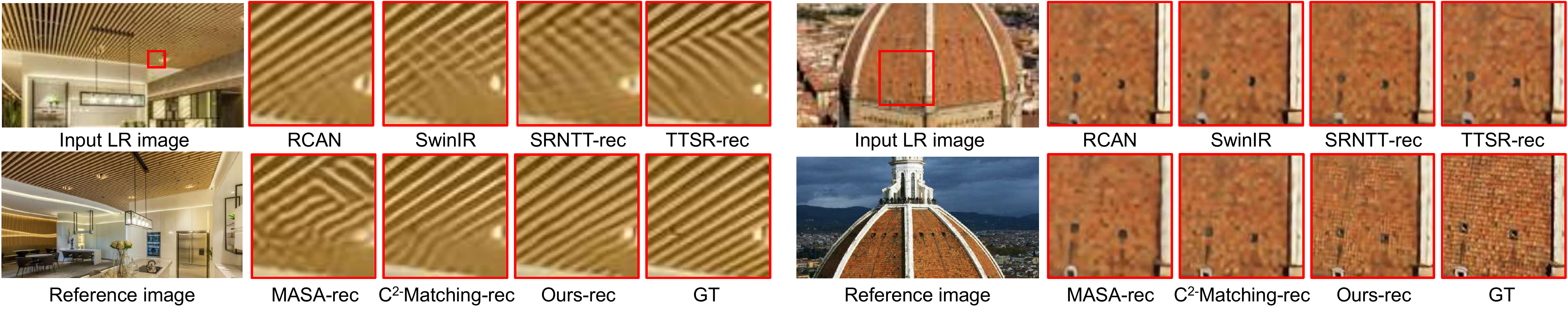}
\caption{Qualitative comparisons of SISR and RefSR models trained with the reconstruction loss.}
\label{fig:qua_comp_rec}
\end{figure*}

\begin{figure*}[t]
\setlength\belowcaptionskip{-10pt}
\centering
\includegraphics[width=1\linewidth,trim=0 0 0 50]{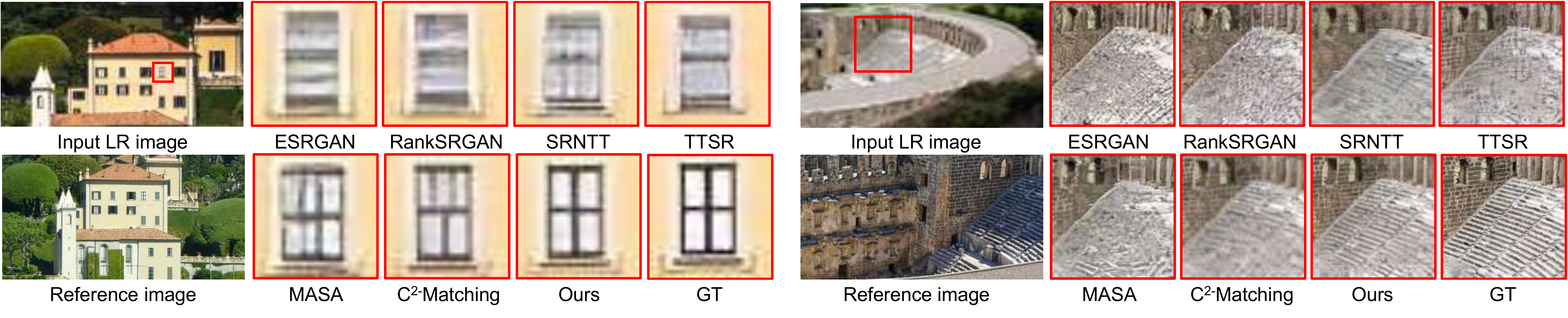}
\includegraphics[width=1\linewidth,trim=0 40 0 0]{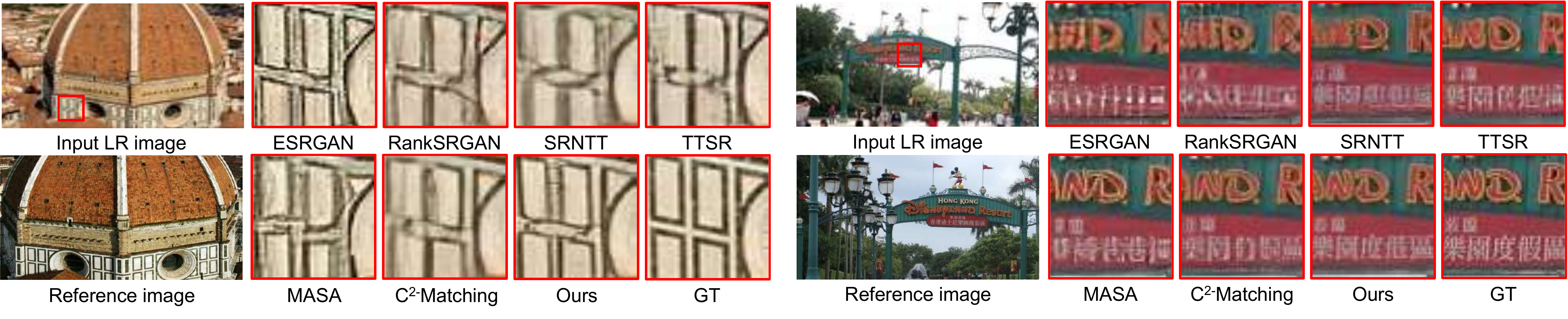}
\caption{Qualitative comparisons of SISR and RefSR models trained with all loss.}
\label{fig:qua_comp_gan}
\end{figure*}

\subsection{Comparison with State-of-the-art Methods}
We compare with the SISR methods (SRCNN \cite{dong2015image}, EDSR \cite{lim2017enhanced}, RCAN \cite{zhang2018image}, SwinIR \cite{liang2021swinir}, SRGAN \cite{ledig2017photo}, ENet \cite{sajjadi2017enhancenet}, ESRGAN \cite{wang2018esrgan}, and RankSR-GAN \cite{zhang2019ranksrgan}) and RefSR methods (CrossNet \cite{zheng2018crossnet},  SRNTT \cite{zhang2019image}, SSEN \cite{Shim_2020_CVPR}, TTSR \cite{yang2020learning}, E2ENT2 \cite{xie2020feature}, and MASA \cite{lu2021masa}). 
For fair comparisons, the above models are trained on CUFED5 training set, and tested on CUFED5 testing set, Urban100, Manga109, Sun80 and WR-SR.
In this experiment, we train our model on two cases only with reconstruction loss (denoted as `-rec'), and with all loss functions.

\noindent{\textbf{{Quantitative comparison.}}}
We provide quantitative comparisons of SR models trained with only reconstruction loss and all losses in Tables \ref{tab:quan_comp_rec} and \ref{tab:quan_comp_gan}, respectively.
In Table \ref{tab:quan_comp_rec}, our model has the best PSNR and SSIM on all testing sets and significantly outperforms all SISR and RefSR models.
It implies that our Transformer achieves the state-of-the-arts and good generalization performance.
For the SISR setting, our method performs better than the state-of-the-art SISR method \cite{liang2021swinir}.
It is difficult for these SISR methods to synthesize   since the high-frequency information is degraded.
In contrast, our model is able to adaptively discover the useful information from a reference image on the Urban100 and Manga109 datasets even if it is a random image. For the RefSR setting, our proposed DATSR significantly outperforms all methods with the help of the cooperative transfer with deformable convolution module.

In Table \ref{tab:quan_comp_gan}, our DATSR also achieves the much higher PSNR/SSIM values than other RefSR methods with a large margin.
Our DATSR trained with adversarial loss reduces PSNR and SSIM but increases the visual quality. Still, it has the best performance over all compared methods.
The above quantitative comparison results on different SR paradigms demonstrate the superiority of our Transformer over state-of-the-art SISR and RefSR methods.

\noindent\textbf{Qualitative comparison.}
The visual results of our method are shown in Figs. \ref{fig:qua_comp_rec} and \ref{fig:qua_comp_gan}.
In these figures, our model also achieves the best performance on visual quality when trained with the reconstruction loss and all loss.
These results demonstrate that our proposed method is able to transfer more accurate textures from the Ref images to generate SR images with higher quality.
When trained with the reconstruction loss, our medel can synthesize SR images with sharp structure.
Moreover, our method is able to search and transfer meaningful texture in a local regions even if the Ref image is not globally relevant to the input image.
When trained with the adversarial loss, our model is able to restore the realistic details in the output images which are very close to the HR ground-truth images with the help of the given Ref images.
In contrast, it is hard for ESRGAN and RankSRGAN to generate realistic images without the Ref images since the degradation is severely destroyed and high frequency details of images are lost.
For RefSR methods, our model is able to synthesize more realistic texture from the Ref images than SRNTT \cite{zhang2019image}, TTSR \cite{yang2020learning}, MASA \cite{lu2021masa}, and $C^2$-Matching \cite{jiang2021robust}.
For example, in the top of Fig. \ref{fig:qua_comp_gan}, our model is able to recover the ``window" with sharper edge and higher quality than $C^2$-Matching, but other methods fail to restore it even if they have a Ref image.

\begin{figure*}[!t]
    \setlength\belowcaptionskip{-5pt}
    \setlength\abovecaptionskip{-1pt}
	\includegraphics[width=1\linewidth,trim=20 0 20 0 ]{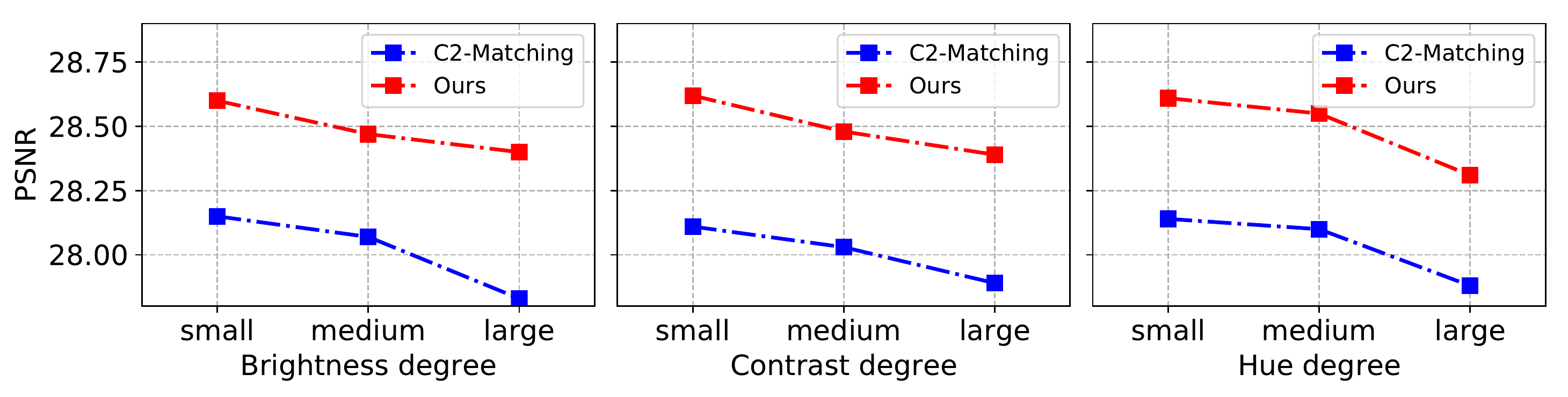}
	\caption{Robustness to different image transformations. Our DATSR is more robust than $C^2$-Matching \cite{jiang2021robust} under different image transformations.}
	\label{fig:image_trans}
\end{figure*}

\begin{figure}[!t]
    \setlength\belowcaptionskip{-5pt}
    \setlength\abovecaptionskip{-1pt}
	\subfigure[Extreme cases for Ref images.]{   
		\begin{minipage}{6cm}
			\centering    
			\includegraphics[scale=0.21,trim=20 0 0 0]{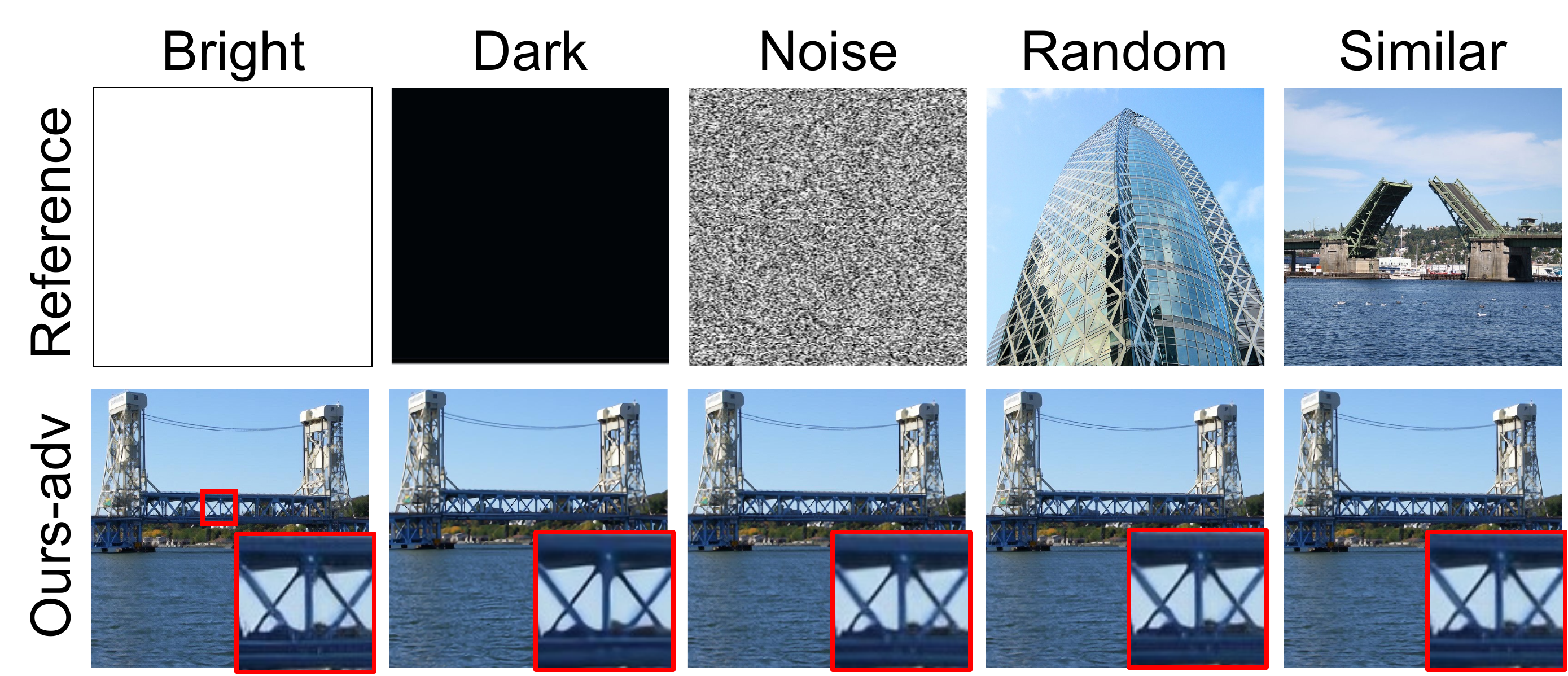} 
		\end{minipage}
	}
	\subfigure[Different sources of Ref images.]{ 
		\begin{minipage}{6cm}
			\centering    
			\includegraphics[scale=0.21,trim=30 0 0 0]{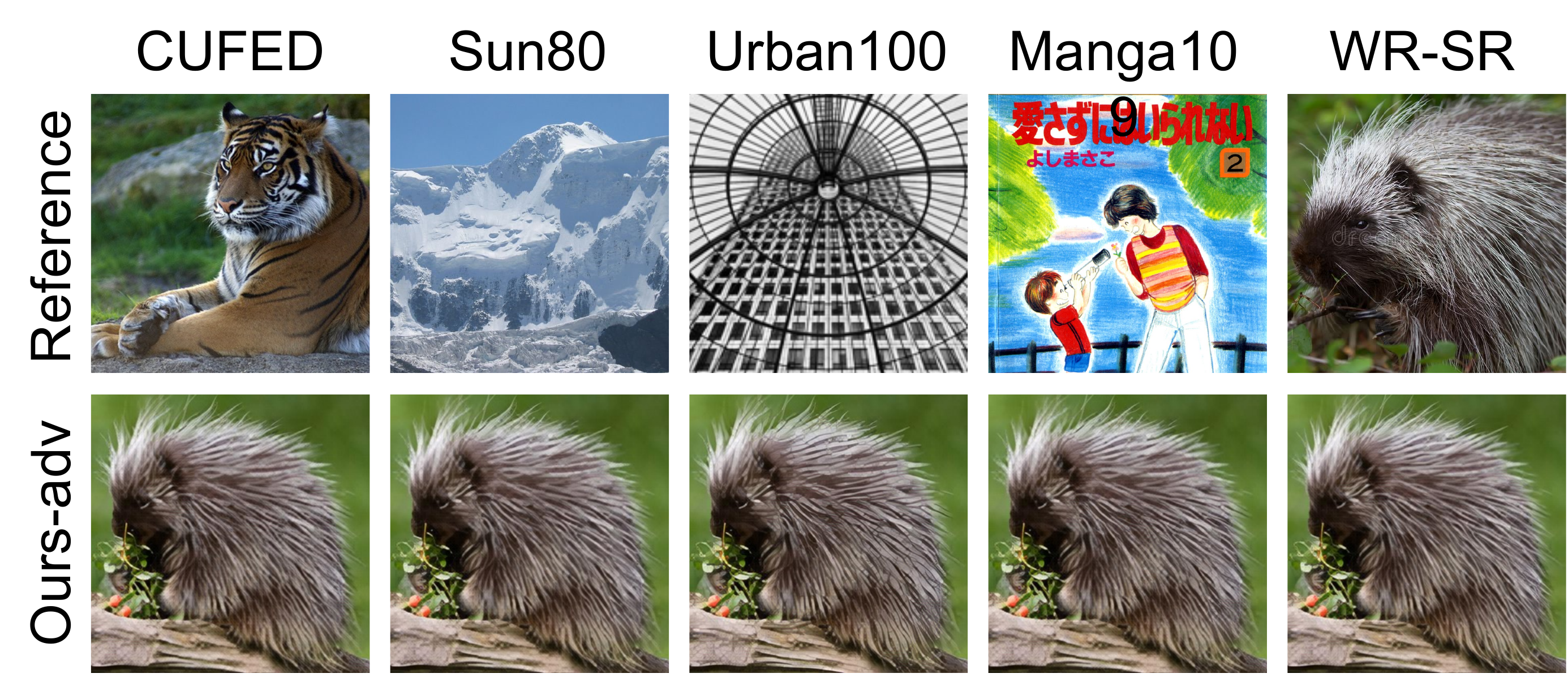}
		\end{minipage}
	}
	\caption{Investigation on different types of reference images.}  
	\label{fig:diffref}    
\end{figure}

\begin{figure}[t]
\setlength\belowcaptionskip{-2pt}
\centering
\begin{minipage}[t]{0.48\textwidth}
\centering
\includegraphics[width=5.5cm,trim=0 40 0 0 ]{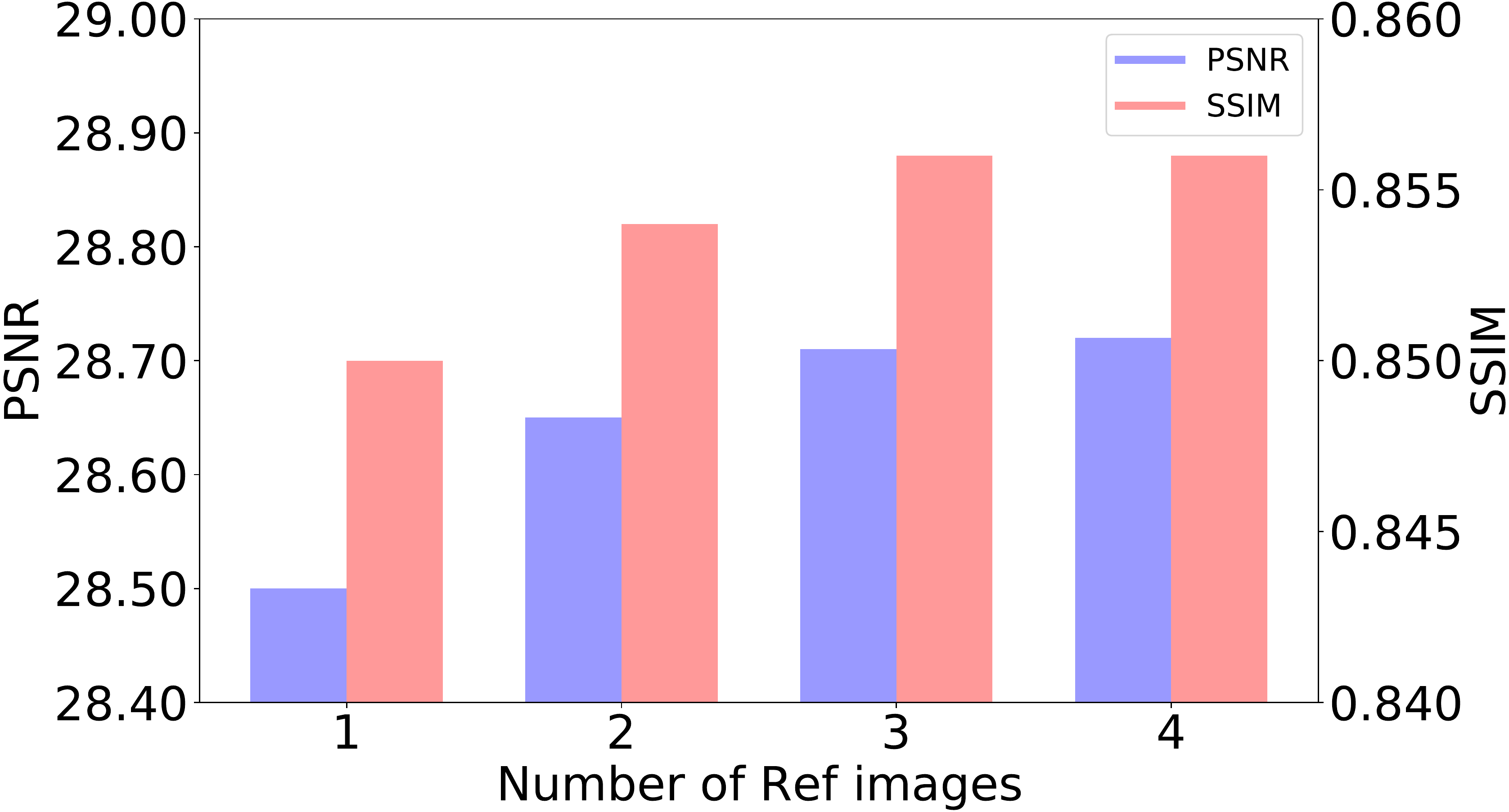}
\caption{Effect on \#Ref images.}
\label{fig:effect_ref_no}
\end{minipage}
~~~~~
\begin{minipage}[t]{0.46\textwidth}
\centering
\includegraphics[width=5cm,trim=0 20 0 0]{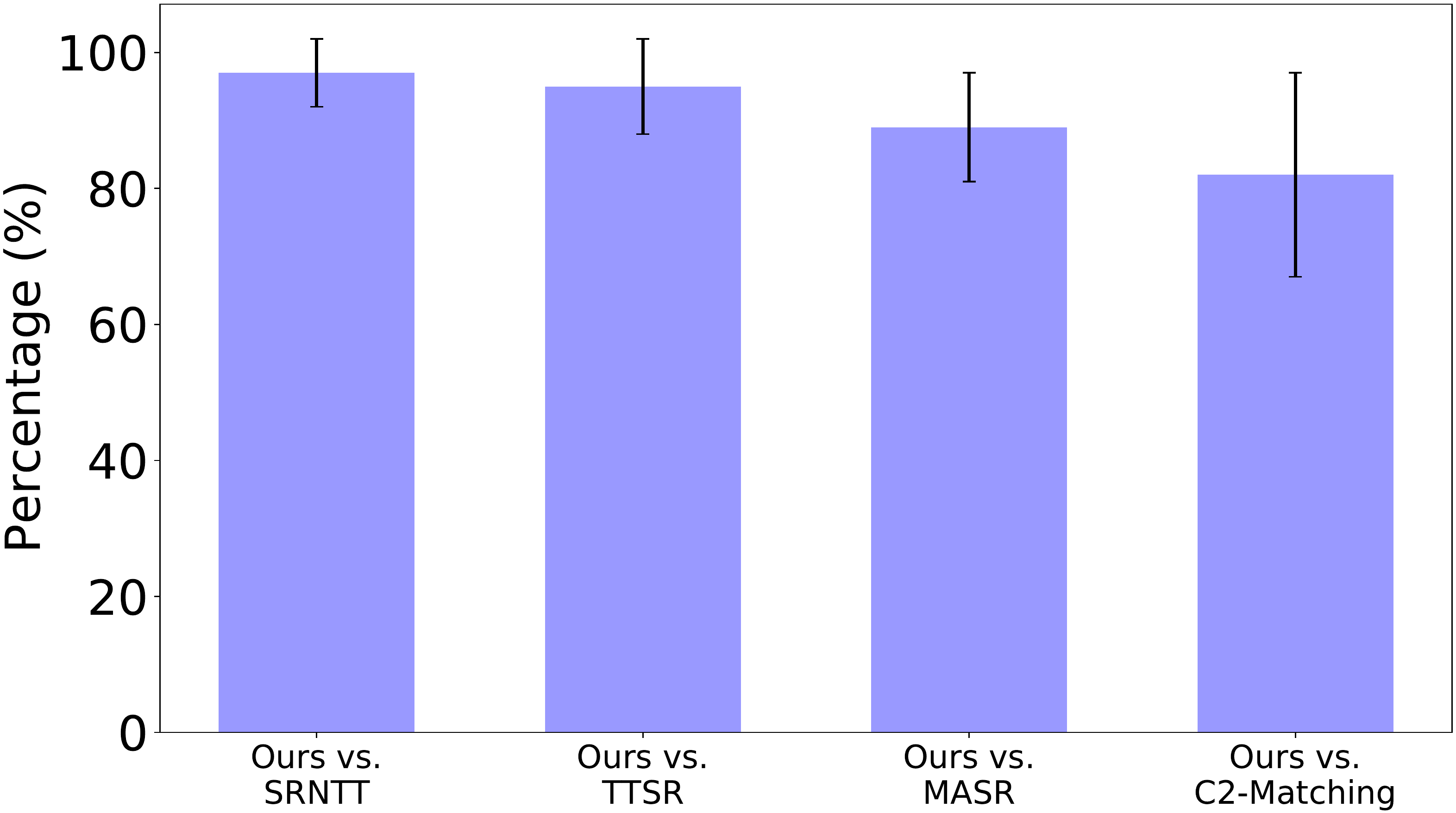}
\caption{User study.}
\label{fig:user_study}
\end{minipage}
\end{figure}

\begin{table}[t]
\setlength\belowcaptionskip{-5pt}
\setlength\abovecaptionskip{-9pt}
\caption{Performance in terms of different similarity levels on CUFED5 test set.}
\begin{center}
\resizebox{1\textwidth}{!}{
\begin{tabular}{|l|c|c|c|c|c|c|c|c|c|c|}
\hline
\multirow{2}{*}{Similarity levels}        & \multicolumn{2}{c|}{L1} & \multicolumn{2}{c|}{L2} & \multicolumn{2}{c|}{L3} & \multicolumn{2}{c|}{L4} & \multicolumn{2}{c|}{Average} \\
\cline{2-11}
& PSNR & SSIM & PSNR & SSIM & PSNR & SSIM & PSNR & SSIM & PSNR & SSIM \\
\hline\hline 
CrossNet \cite{zheng2018crossnet} & 25.48 & 0.764 & 25.48 & 0.764 & 25.47 & 0.763 & 25.46 & 0.763 & 25.47 & 0.764 \\
SRNTT-rec \cite{zhang2019image} & 26.15 & 0.781 & 26.04 & 0.776 & 25.98 & 0.775 & 25.95 & 0.774 & 26.03 & 0.777 \\
TTSR-rec \cite{yang2020learning} & 26.99 & 0.800 & 26.74 & 0.791 & 26.64 & 0.788 & 26.58 & 0.787 & 26.74 & 0.792 \\
$C^2$-Matching-rec \cite{jiang2021robust} & 28.11 & 0.839 & 27.26 & 0.811 & 27.07 & 0.804 & 26.85 & 0.796 & 27.32 & 0.813 \\
\hline
\bf{DATSR-rec (Ours)} & \bf{28.50} & \bf{0.850} & \bf{27.47} & \bf{0.820} & \bf{27.22} & \bf{0.811} & \bf{26.96} & \bf{0.803} & \bf{27.54} & \bf{0.821} \\
\hline
\end{tabular}
}
\label{tab:diff_sim}
\end{center}
\end{table}

\subsection{Further Analyses}
\noindent\textbf{Robustness to image transformations.}
We analyze the robustness of our model to different kinds of image transformations.
Specifically, we use ColorJitter to augment the CUFED5 testing set by randomly change the brightness, contrast and hue of Ref images into three group: small, medium and large. 
The detailed settings are put in the supplementary materials.
In Fig. \ref{fig:image_trans}, our model is more robust than $C^2$-Matching \cite{jiang2021robust} under different image transformations.
Note that the medium and large transformations are not included during training but our model still has superior performance.

\paragraph{\textbf{\emph{Effect on type and number of Ref images.}}}
We test our model on different Ref images, such as extreme images (\ie may have only one color or noise without any information) and random images from different testing sets.
In Fig. \ref{fig:diffref}, our method has robust performance and high visual quality even if the Ref images have no useful texture information.
In addition, our model has better performance when increasing \#Ref images in Fig.~\ref{fig:effect_ref_no}. Table \ref{tab:diff_sim} shows the results of four similarity levels (``L1'' to ``L4'') where L1 is the most relevant level.
Our method achieves the best performance across all similarity levels.

\paragraph{\textbf{\emph{Comparisons with multi-RefSR methods.}}}
We compare our model with multi-RefSR methods, \ie CIMR-SR \cite{yantowards} and AMRSR \cite{pesavento2021attention}.
Note that these multi-RefSR methods are trained with a collection of reference images.
In Table \ref{tab:quan_comp_multi_refsr}, our model trained with single reference image performs better than CIMR-SR and AMRSR with many reference images, which further demonstrate the superiority of our proposed DATSR.

\begin{figure}[t]
	\centering  
	\begin{minipage}[b]{0.59\linewidth}
		\setlength\abovecaptionskip{-0.5pt}
		\captionof{table}{Comparisons with Multi-RefSR on the CUFED5 testing set.}
		\label{tab:quan_comp_multi_refsr}
		\centering    
		\resizebox{1\textwidth}{!}{
			\begin{tabular}{|l|c|c|c|}
            \hline
            Methods   & CIMR-SR \cite{yantowards} & AMRSR \cite{pesavento2021attention} & \bf{DATSR-rec} \\ \hline\hline
            w/ rec. loss  & 26.35/0.789 & 28.32/0.839 & \textbf{28.72}/\textbf{0.856} \\
            w/ all losses & 26.16/0.781 & 27.49/0.815 & \textbf{27.95}/\textbf{0.835} \\
            \hline
            \end{tabular}
		}
	\end{minipage}
	~
	\begin{minipage}[b]{.38\linewidth}
	    \setlength\belowcaptionskip{-5pt}
        \setlength\abovecaptionskip{-0.1pt}
		\captionof{table}{Comparisons of LPIPS~\cite{zhang2018unreasonable} with $C^{2}$-Matching.}
		\label{table_lpips}
		\centering    
		\resizebox{1\textwidth}{!}{
			\begin{tabular}{|l|c|c|}
				\hline
				{Methods}             & {CUFED5} & {WR-SR} \\ \hline\hline 
				$C^{2}$-Matching \cite{jiang2021robust} & 0.164 & 0.219 \\ 
				\bf{DATSR (Ours)} & \bf{0.140} & \bf{0.211} \\
				\hline
			\end{tabular}
		}
	\end{minipage}
\end{figure}

\begin{figure}[t]
	\centering  
	    \begin{minipage}[b]{.48\linewidth}
	    \setlength\belowcaptionskip{-5pt}
        \setlength\abovecaptionskip{-0.1pt}
		\captionof{table}{Comparisons of model size and performance with $C^2$-Matching.}
		\label{table_param_comp}
		\centering    
		\resizebox{1\textwidth}{!}{
			\begin{tabular}{|l|c|c|c|}
				\hline
				{Methods}  & Params & PSNR & SSIM \\ \hline \hline
				TTSR-rec \cite{yang2020learning} & 6.4M & 27.09 & 0.804 \\ 
				$C^{2}$-Matching-rec \cite{jiang2021robust} & 8.9M & 28.24 & 0.841 \\ 
				\bf{DATSR-rec (Ours)} & 18.0M & \bf{28.72} & \bf{0.856} \\
				\hline
			\end{tabular}
		}
	\end{minipage}
		~~
		\begin{minipage}[b]{.46\linewidth}
		    \setlength\abovecaptionskip{-0.5pt}
			\captionof{table}{Ablation study on the RDA and RFA modules.}
			\label{table:ablation}
			\centering    
			\resizebox{1\textwidth}{!}{
				\begin{tabular}{|l|c|c|}
					\hline
					Methods      & PSNR & SSIM \\ \hline\hline 
					RDA (w/ feature warping)~~ & 28.25 & 0.844 \\
					RFA (w/ ResNet blocks) & 28.50 & 0.850  \\
					\bf{DATSR-rec}        & \bf{28.72} & \bf{0.856}  \\
					\hline
				\end{tabular}
			}
		\end{minipage}
\end{figure}

\subsection{More Evaluation Results}
\noindent\textbf{Perceptual metric.}
We further use the perceptual metric LPIPS \cite{zhang2018unreasonable} to evaluate the visual quality of the generated SR images on the CUFED5 and WR-SR testing sets.
Recently, this metric is also widely used in many methods \cite{lucas2019generative,lugmayr2020ntire}.
In general, smaller LPIPS corresponds to the better performance for RefSR.
As shown in Table \ref{table_lpips}, our model achieves smaller LPIPS than $C^2$-Matching.
Thus, our model generates SR images with better quality than $C^2$-Matching.

\paragraph{\textbf{\emph{User study.}}}
To further evaluate the visual quality of the SR images, we conduct the user study to compare our proposed method with previous state-of-the-art methods, including SRNTT \cite{zhang2019image}, TTSR \cite{yang2020learning}, MASA \cite{lu2021masa} and $C^2$-Matching \cite{jiang2021robust} on the WR-SR testing set.
The user study contains 20 users, and each user is given multiple pairs of SR images where one is our result.
Then, each user chooses one image with better visual quality.
The final percentage is the average user preference of all images.
In Fig.~\ref{fig:user_study}, over 80\% of the users prefer that our results have better quality than existing RefSR methods.

\subsection{Discussion on Model Size}
To further demonstrate the effectiveness of our model, we also show the comparison of model size (\ie the number of trainable parameters) with the state-of-the-art model (\ie $C^2$-Matching \cite{jiang2021robust}) in Table~\ref{table_param_comp}.
Our model has a total number of 18.0M parameters and achieves PSNR and SSIM of 28.72 and 0.856, respectively.
The results demonstrate that our proposed model outperforms $C^2$-Matching with a large margin, although our model size is higher than this method.
The part of our model size comes from the Swin Transformer in the RFA module.
More discussions of other RefSR models are put in the supplementary materials.

\subsection{Ablation Study}

We first investigate the effectiveness of RDA and RFA in Table \ref{table:ablation}.
Specifically, we replace the texture transfer method in RDA with a feature warping based on the most relevant correspondence, and replace RFA with several convolutional neural networks (CNNs).
The model with feature warping or CNNs is worse than original model with RDA or RFA.
Therefore, RDA is able to discover more relevant features especially when the correspondence is not inaccurate. 

For RFA, our model has better performance than the directly using simple CNNs.
Nevertheless, with the help of RDA, training with CNNs still outperforms $C^2$-Matching with large margin. 
Therefore, it verifies that the effectiveness of RFA and it is able to aggregate the features at different scales. 
More discussions on ablation studies are put in the supplementary materials.

\section{Conclusion}
In this work, we propose a novel reference-based image super-resolution with deformable attention Transformer, called DATSR.
Specifically, we use texture feature encoders module to extract multi-scale features and alleviate the resolution and transformation gap between LR and Ref images.
Then, we propose reference-based deformable attention module to discover relevant textures, adaptively transfer the textures, and relieve the correspondence mismatching issue. 
Last, we propose a residual feature aggregation module to fuse features and generate SR images.
Extensive experiments verify that DATSR achieves the state-of-the-arts performance as it is robust to different brightness, contrast, and color between LR and Ref images, and still shows good robustness even in some extreme cases, when the Ref images have no useful texture information.
Moreover, DATSR trained with a single Ref image has better performance than existing Multi-RefSR methods trained with multiple Ref images.

\paragraph{\textbf{\emph{Acknowledgements.}}}
This work was partly supported by Huawei Fund and the ETH Zürich Fund (OK).
%
%
\bibliographystyle{splncs04}
\bibliography{egbib}
\end{document}